\RequirePackage{fix-cm}
\documentclass[smallcondensed]{svjour3}     
\smartqed  
\usepackage{graphicx}
\usepackage{caption}
\usepackage{url}
\usepackage{longtable}
\usepackage{multirow}   
\usepackage{amssymb}

\begin{document}

\title{Understanding Substructures in Commonsense Relations in ConceptNet 
}


\author{Ke Shen         \and
        Mayank Kejriwal 
}


\institute{M. Kejriwal (Corresponding Author), K. Shen \at
              Information Sciences Institute, University of Southern California \\
              4676 Admiralty Way \\
              Marina del Rey, California 90292 \\
              United States of America
              Tel.: +310-822-1511\\
              Fax: +1310-823-6714\\
              \email{kejriwal@isi.edu}           
}

\date{Received: date / Accepted: date}

\maketitle

\begin{abstract}
Acquiring commonsense knowledge and reasoning is an important goal in modern NLP research. Despite much progress, there is still a lack of understanding (especially at scale) of the nature of commonsense knowledge itself. A potential source of structured commonsense knowledge that could be used to derive insights is ConceptNet. In particular, ConceptNet contains several coarse-grained relations, including `HasContext', `FormOf' and `SymbolOf', which can prove invaluable in understanding broad, but critically important, commonsense notions such as `context'. In this article, we present a methodology based on unsupervised knowledge graph representation learning and clustering to reveal and study substructures in three heavily used commonsense relations in ConceptNet. Our results show that, despite having an `official' definition in ConceptNet, many of these commonsense relations exhibit considerable \emph{sub-structure}. In the future, therefore, such relations could be sub-divided into other relations with more refined definitions. We also supplement our core study with visualizations and qualitative analyses.
\end{abstract}

\section{Introduction}\label{introduction}

Despite the ubiquity of intelligent agents such as Alexa and Siri in modern life, these agents have yet to capture the human element in natural conversations. Even with advances in Natural Language Processing (NLP), deep learning, and knowledge graphs  \cite{Hirschberg261}, \cite{QGuilin}, it is not clear if such agents are fully capable of answering questions (e.g., `Should I put my spare change in a piggy bank?') with incomplete information, or under-specified needs. Such questions tend to require more contextual and implicit knowledge that humans often take for granted when navigating daily situations. Among other things, lack of deep contextual understanding limits the agents' \emph{commonsense reasoning} abilities.

Commonsense reasoning is the process that involves processing information about a scenario in the world, and making inferences and decisions by using context, implicit and explicit information based on our collective `commonsense knowledge'. Commonsense knowledge is difficult to define precisely but it is usually assumed to be a broad body of knowledge of how the `world' works \cite{Mueller}. Generally, such knowledge is essential for navigating social situations and interactions, `naive' physical understanding (e.g., the simple knowledge that when an object on the table is `picked up', it is not on the table anymore) and more controversially, knowledge that relies on reasoning about local culture and milieu \cite{socialiqa}, \cite{Cyc}. 

Commonsense knowledge and reasoning have both been recognized as essential for building more advanced `general' AI systems that have human-like capabilities and reasoning ability when facing uncertain, implicit, or even potentially contradictory, information. Recognizing its importance, researchers in several communities have increasingly engaged in improving agent performance on commonsense question answering, abductive reasoning and other tasks pertinent to commonsense reasoning \cite{Davis}, \cite{Melissa}, \cite{socialiqa}, \cite{sap-etal-2020-commonsense}.

\begin{figure}[h]
  \centering
  \includegraphics[width=12cm]{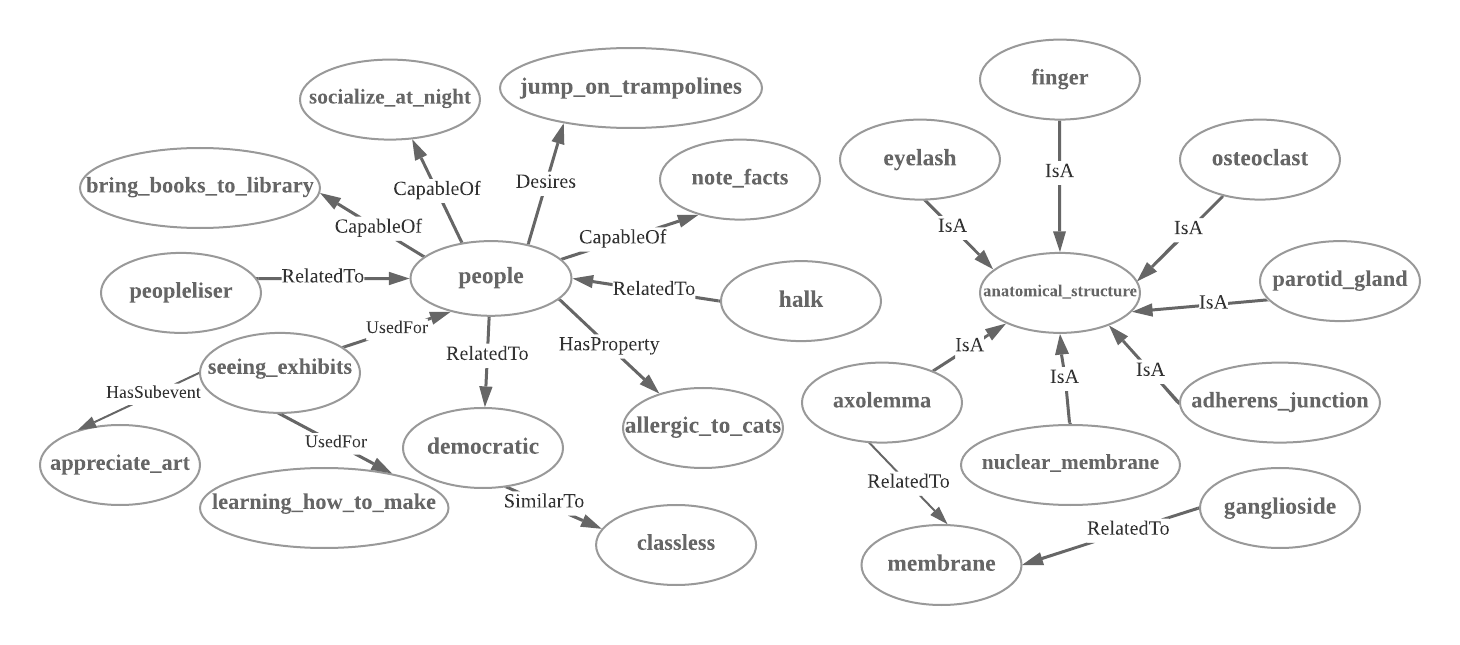}
  \caption{A fragment of the ConceptNet Knowledge Graph (KG).}
  \label{figure: conNet_frag}
\end{figure}

ConceptNet\footnote{\url{https://conceptnet.io/}.} is a large-scale, freely available knowledge graph (KG) that describes commonsense knowledge as a set of assertions or \emph{triples} \cite{Conceptnet5}. It is designed to represent the common knowledge needed to help machines better understand the meanings of concepts and inter-concept relationships that people rely on in everyday situations. The \emph{graph structure} that represents knowledge in ConceptNet is particularly useful for textual reasoning over natural language documents. An example of how such knowledge is organized in ConceptNet is illustrated in Figure \ref{figure: conNet_frag}. ConceptNet originated from the Open Mind Common Sense \cite{OMCS} project, itself launched in 1999 at the MIT Media Lab. It was regularly updated to include new knowledge from crowdsourced resources, expert-curated resources, and `games with a purpose' designed specifically to elicit certain kinds of commonsense annotations from people, such as intuitive word associations. 

Our guiding hypothesis in this paper is that, due to its growth and usage over the years, ConceptNet can yield valuable insights about commonsense knowledge itself. The intuition behind our approach is relatively straightforward. First, using representation learning, we `embed' each entity and relation into a dense, continuous real-valued vector space, learned in an unsupervised fashion using a state-of-the-art knowledge graph representation learning algorithm. Next, we isolate the \emph{triples} [or labeled edges, such as (people, CapableOf, socialize\_at\_night) in Figure \ref{figure: conNet_frag}] of interest from the raw knowledge base. For example, we isolate the triples with relation `hasContext' if we want to study that relation. Finally, using the learned embeddings from the first step, we derive an embedding for the triple using the notion of \emph{translation} that is more formally described subsequently. Once such embeddings have been obtained for each triple with the relation under study, we cluster them using an established algorithm such as k-Means to detect potential sub-structures. We also use statistical and quantitative measures to understand the quality and structural properties of these clusters. Finally, visualization and sampling-based qualitative analyses are used to provide more insight into the data. Specific contributions are enumerated below:

\begin{itemize}
\item We present a novel methodology for understanding structural aspects of commonsense knowledge by studying three relations (`HasContext', `SymbolOf', and `FormOf'). These relations are both \emph{high-volume} and \emph{coarse-grained}, and are known to be significant in commonsense reasoning, explaining their high prevalence in ConceptNet.  Our methodology relies on a judicious combination of knowledge graph representation learning, clustering and visualization.
\item We conduct a detailed set of experiments by applying the proposed methodology  on a large sample of triples that contain the three relations mentioned earlier. Using quantitative metrics for measuring cohesion and separation in the embedding space, our results show that, despite having an `official' definition in ConceptNet, many of these high-volume, coarse-grained relations exhibit considerable \emph{sub-structure}. In the future, therefore, such relations could be `divided' into other relations with more refined definitions.
\item   Using qualitative and sampling-based analyses, we show how a broad notion like `context' is defined in practice in commonsense knowledge bases such as ConceptNet. These analyses potentially allow us to develop a deeper understanding of the definition and role of  context in commonsense knowledge.
\end{itemize}

While our goal is not to present a complete understanding of common sense,  we study the three relations noted earlier (`HasContext', `SymbolOf', and `FormOf') to understand relations that are believed to be critical to common sense reasoning. Our methodology and empirical study also relies on established methods from the KG representation learning and clustering literature. An explicit goal behind using established tools, besides their expected reliability, is to enable greater replicability for studying other such relations, including in other knowledge bases.

The rest of this work is structured as follows. In Section 2, we describe relevant lines of research related to this work. In Section 3, we present our materials and methods, followed by the experimental results in Section 4. Section 5 discusses the key implications. The article concludes in Section 6.

\section{Related Work}\label{relatedwork}

Although both Wikipedia\footnote{\url{https://en.wikipedia.org/wiki/Common_sense}.} and the (relatively) recent Defense Advanced Research Projects Agency (DARPA) Machine Common Sense (MCS) program\footnote{\url{https://www.darpa.mil/program/machine-common-sense}.} define commonsense reasoning as `the basic ability to perceive, understand, and judge things that are shared by nearly all people and can be reasonably expected of nearly all people without need for debate', there is no official, sufficiently broad definition that we can cite outside of the psychology context. Within psychology, \cite{smedslund1} has defined commonsense as `the system of implications shared by the competent users of a language'. One important commonality that is shared, no matter the definition or field considered, is that commonsense knowledge is (at least to a degree) \emph{implicit}. 

%

There is very little work on the \emph{nature} of commonsense knowledge: one rare example is a recent book \cite{csbook1} that attempts to provide a `theory' of commonsense by breaking down commonsense knowledge into various categories, and present formalisms for those categories. Other similar work along those lines include \cite{csaxioms1}, \cite{csaxioms2}, \cite{csaxioms3}. Unlike those works, we take an inductive, data-driven approach. Our work is potentially complementary to \cite{csbook1}, since some of our findings may be used in the future to provide evidence for (or against) some of their purely theoretical claims.

Progress has been achieved in specific \emph{kinds} of commonsense reasoning, especially in reasoning about time and internal relations \cite{taxonomy1}, \cite{Pinto}, reasoning about actions and change \cite{Srinivas}, and the sign calculus \cite{Davis}. Semantics have played an important role in some of these successes \cite{semantics1}; in fact, ConceptNet itself has been described as a `semantic network' \cite{conceptnet3}. A similar example is WordNet, whose entries are organized in terms of semantic inter-relationships. The easy-to-use network structure lends it to being frequently applied in multiple reasoning systems \cite{botschen}, \cite{angeli}, \cite{Lin2017}. Other relevant areas in AI that could potentially benefit from this work include graph-based models \cite{sncs1}, meta-heuristics \cite{sncs3}, sarcasm detection \cite{sncs2}. 

%

Commonsense reasoning systems are typically measured against benchmark tests, some well known examples of such datasets including abductive Natural Language Inference (aNLI) \cite{anli}, HellaSwag \cite{hellaswag} and Physical Interaction Question Answering (PIQA) \cite{PIQA}. However, in this article, our focus is on understanding the nature of commonsense by studying a knowledge base such as ConceptNet, rather than building a reasoning system that performs well on benchmarks. We note that our findings can potentially be used to enhance performance of such systems. Recent work that combines knowledge bases with language representation learning models to improve commonsense question answering performance has been promising.

While knowledge bases and knowledge graphs have been extensively used for representing encyclopedic knowledge and other domains \cite{JensDbpedia}, \cite{wordnet}, \cite{freebase}, ConceptNet  and Cyc are the only two examples that we are aware of that serve as commonsense KBs \cite{conceptnet5.5}, \cite{Cyc}. Owned by Cycorp, Cyc's knowledge base claims\footnote{\url{https://www.cyc.com/}.} to be the `broadest, deepest, and most complete repository ever developed', but given its proprietary nature, the claim is difficult to validate. It is not known how similar the commonsense content of Cyc is to that of ConceptNet, but potentially, the same methodology proposed in this paper could be applied to Cyc to replicate, strengthen or refute some of our findings. ConceptNet also expresses knowledge in an easy-to-use format (i.e., as sets of 3-tuple assertions, illustrated as edges in Figure \ref{figure: conNet_frag}) rather than in higher-order logic notation. Because of these advantages, ConceptNet has rapidly emerged as a practical dataset and resource for various kinds of machine learning and NLP in the last ten years in particular \cite{CNapp1}, \cite{CNapp2}, \cite{CNapp3}.

Compared with \emph{WordNet} \cite{wordnet}, which focuses on preserving lexicographic information and the relationship between words and their senses, ConceptNet maintains a semantic network structure that is designed to capture commonsense assertions. In particular, ConceptNet contains more relations than WordNet.

In contrast to Cyc, ConceptNet is a freely available multilingual KG that connects everyday entities using a rich set of relations. As mentioned in the introduction, ConceptNet serves as an important background resource for commonsense reasoning and question answering. However, it has not been studied directly for gaining insights into commonsense knowledge, even though there is precedent. For example, studies on DBpedia and YAGO have been conducted specifically to understand their relational structure and the structural properties of the encyclopedic knowledge that these KGs are known for \cite{DBpediaStudy},\cite{YAGOStudy}, \cite{KGComp}. We attempt to do the same, but with commonsense knowledge as the focus.  

Another line of work highly related to this one is \emph{representation learning}, especially as they pertain to KGs. Such algorithms take as input the KG, including entities and relations, and embed them into continuous vector spaces, as surveyed by \cite{QWang}. Models such as RESCAL \cite{RESCAL}, TransE \cite{transE}, TransH \cite{transh}, KG2E \cite{KG2E}, and RotateE \cite{rotateE} all achieve good performance on various tasks, such as KG completion \cite{transE} and relation extraction \cite{JWeston}, which allows for further improvement of the KG. In this article, we use a graph embedding package that builds on the ones above (especially TransE) and is especially designed for graphs with many millions of triples. 


Finally, we note recent advances in commonsense reasoning and question answering by using transformer-based language representation learning models such as Bidirectional Encoder Representations from Transformers (BERT) \cite{bert}, GPT-3 \cite{gpt3}, UnifiedQA \cite{unifiedqa}, and several others. There is also a trend in improving zero-shot learning with commonsense reasoning \cite{zscs1}, \cite{zscs2}. Separately, ensemble applications of symbolic and sub-symbolic AI models have also been proposed for commonsense reasoning \cite{ensemble1}, \cite{ensemble2}. While these advances have led to state-of-the-art performance across NLP tasks, of which question answering is only one example, these models do not help us understand the nature of commonsense reasoning itself. Developing such an understanding is necessary both as a scientific mode of investigation into human commonsense (which is ubiquitous), and to further improve transformers and other neural models to be more \emph{interpretable} when producing answers to questions. This work does not propose a new method for achieving higher task-specific performance on commonsense benchmarks, but rather, proposes a computational methodology for studying structured commonsense knowledge. By applying a data-driven and empirically grounded methodology on a dataset like ConceptNet, our hope is that we can learn more about commonsense as a fundamental phenomenon. 


\section{Materials and Methods}

Our guiding principle in this paper is that a commonsense KB such as ConceptNet could be used as the basis for understanding some of the structural properties of commonsense knowledge. 
Conceptually, ConceptNet can be defined as a multi-relational graph $G = (V, R, E)$, where V is the set of entities or \emph{nodes}, R is the set of 49 \emph{relations} and E is the set of triples or \emph{edges} where each triple $e = (h, r, t) \in E \subseteq V \times R \times V$. While $G$ can also be thought of in a non graph-theoretic way as a \emph{set} of triples\footnote{This definition is sometimes used in the early natural language community when discussing these datasets as \emph{knowledge bases} rather than (multi-relational) \emph{knowledge graphs}, where triples are usually interpreted as directed edges \cite{kejriwalDSKG}.}, the distinction is not relevant for the purposes of this paper, and we use knowledge bases and graphs interchangeably.  However, it is more natural to think about ConceptNet as a graph due to its structural properties. Earlier, Figure  \ref{figure: conNet_frag} expressed a fragment of ConceptNet as a multi-relational graph with 21 edges, or alternatively, 21 triples. Another example-set of actual triples from ConceptNet is also provided in Table \ref{triplesexamples}.

\begin{table}[h]
\centering
\begin{tabular}{|c|}

\hline
{\bf Example triple}                                      \\ \hline
(/c/en/appear/ , /r/Antonym/ , /c/en/hide/) \\ \hline
(/c/en/apparent/a/ , /r/Antonym/ , /c/en/inapparent/) \\ \hline
(/c/en/jury , /r/CapableOf , /c/en/state\_verdict) \\ \hline
(/c/en/accriminate , /r/DerivedFrom , /c/en/criminate/v) \\ \hline
(/c/en/mutton\_ham/n , /r/RelatedTo , /c/en/salt) \\ \hline

\end{tabular}
\caption{Five actual example triples (each using a different relation) from ConceptNet}
\label{triplesexamples}
\end{table}

The symbols, $h$ and $t$, are respectively referred to as the \emph{source} (or the \emph{head}) and the \emph{destination} (or the \emph{tail}) entity, and $r$ is the relation. For the purposes of maintaining consistent terminology, we use the terms \emph{triple}, \emph{head}, \emph{tail}, and \emph{relation} to refer to the elements $e$, $h$, $t$ and $r$ respectively. Where applicable, these head and tail entities are collectively referred to as \emph{entities}.

Entities and relations are projected into a continuous vector space by PyTorch-BigGraph (PBG) for further computation and analysis \cite{pbg}. Next, we briefly introduce the PBG system. We then describe the raw data and our sampling methodology, followed by the setup and training of the PBG system on our sample.

\subsection{PyTorch-BigGraph (PBG) System}

PBG is an efficient and recent embedding system for learning real-valued vector representations of nodes and edges (referred to as `entities' and `relations' in the context of KGs like ConceptNet) in large-scale graphs. It was developed and publicly released\footnote{\url{https://github.com/facebookresearch/PyTorch-BigGraph}} by Facebook AI Research \cite{pbg}. It is able to deal with large-scale graphs because it relies on distributed computing, in addition to other high-scale techniques, such as graph partitioning and batched negative sampling. At present, it also supports GPU training. 

PBG operates by first training on an input graph, which can be a knowledge graph with labeled edges or `relations'. Similar to other knowledge graph representation learning algorithms (discussed also in the \emph{Related Work}) like TransE and RESCAL  \cite{transE}, \cite{RESCAL}, it outputs embeddings by optimizing an objective, whereby unconnected entities are pushed further apart in the vector space, and entities that are `adjacent' (connected via an edge) are pushed closer to each other. 

Compared to network embeddings such as DeepWalk and node2vec \cite{deepwalk}, \cite{node2vec}, PBG supports multi-entity and multi-relation graph embeddings. Its embedding quality has been found to be comparable with (or even exceed) existing KG embedding systems, evaluated on the Freebase \cite{freebase}, LiveJournal \cite{Livejournal} and YouTube  \cite{youtube} graphs.

\subsection{Raw Data, Sampling and Representation Learning} 

We use a recently released version\footnote{Downloaded at \url{https://github.com/commonsense/conceptnet5/wiki/Downloads}.}, ConceptNet 5.7, for the empirical study in this paper. 
One important aspect of the dataset is the ratio of the number of unique entities to the total number of triples, which is much higher in ConceptNet (0.833) than in other similarly-sized KGs such as Freebase (0.055) or WordNet\footnote{For the interested reader, we are specifically referring to the FB15k and WN18 datasets, as designated in multiple papers on knowledge graph embeddings \cite{SongHJ}, \cite{WZhang}, \cite{FZhao}.} (0.289). Additionally, while ConceptNet tends to contain more entities than `encyclopedic' KGs such as Freebase or DBpedia, there are fewer unique relations.

In practice, these significant deviations in expected entity/relation ratios can cause problems for KG representation learning packages, even those designed for large-scale KGs such as the PBG system. For example, while embedding the full ConceptNet knowledge base, we found that, during the training process, the PBG algorithm fails with a `bus error' message if the number of input triples exceeds 4 million. This occurs due to the algorithm running out of shared memory (despite the fact that we execute the algorithm on a machine with $60+$ GB memory).  To address the memory issue and ensure that our results can be extended or replicated in the future using reasonable computation resources, we randomly sampled 4 million triples from ConceptNet for this study. Key statistics are tabulated in Table \ref{table: triple_statistics}. The head entities-set is twice the size of the tail entities-set and their overlap is approximately 1/20 of the total entities. Cursory analysis also showed that the head entity `/c/en/person' and tail entity `/c/fr/francais' were found to have the most triples associated with them. Other relation-specific statistics are tabulated in Table \ref{table: statistics of triple and entity}. `/r/RelatedTo' was found to be the most frequent relation, occurring in more than 1 million triples.
\begin{table}[h]\scriptsize
\begin{tabular}{|c|c|c|c|c|}

\hline
{\bf Num. triples}    & {\bf Num. entities}    &{\bf Num. head entities} &{\bf Num. tail entities} &{\bf Overlap} \\ \hline
4,000,000 & 3,933,840 & 2,781,892    & 1,387,571    & 235,623                                   \\ \hline

\end{tabular}
\caption{The numbers of triples and entities, including separate head and tail entity counts, and their \emph{overlap} (the number of entities that serve both as head entity or tail entity, possibly in different triples, in the dataset), in the sample of ConceptNet considered in this study.}
\label{table: triple_statistics}
\end{table}

\begin{table}[!h]
\centering
\tiny
\begin{tabular}{|c|p{1.0cm}|p{0.9cm}|c|p{1.0cm}|p{0.9cm}|}
         \hline
{\bf Relation}               &{\bf  Num. triples} &{\bf  Num. entities} &{\bf Relation}               &{\bf  Num. triples} &{\bf  Num. entities} \\ \hline
LocatedNear               & 13          & 26         & dbpedia/leader          & 13         & 19          \\ \hline
CreatedBy                 & 14          & 27          & NotHasProperty          & 44          & 81          \\ \hline
NotCapableOf              & 72          & 132         & dbpedia/capital         & 72          & 137         \\ \hline
Entails                   & 73          & 134         & dbpedia/product         & 81          & 140         \\ \hline
dbpedia/knownFor          & 87          & 168         & dbpedia/field           & 114         & 163         \\ \hline
dbpedia/language          & 151         & 181         & dbpedia/occupation      & 183         & 236         \\ \hline
dbpedia/influencedBy      & 210         & 246         & InstanceOf              & 415         & 570         \\ \hline
DefinedAs                 & 433         & 812         & dbpedia/genus           & 464         & 821         \\ \hline
NotUsedFor                & 519         & 833         & HasLastSubevent         & 571         & 867         \\ \hline
dbpedia/genre             & 621         & 759         & ObstructedBy            & 869         & 1,555       \\ \hline
ReceivesAction            & 988         & 1,721       & CausesDesire            & 1,003       & 1,480       \\ \hline
CapableOf                 & 2,146       & 3,348       & MannerOf                & 2,164       & 2,923       \\ \hline
Antonym                   & 2,601       & 5,100       & HasFirstSubevent        & 2,625       & 3,267       \\ \hline
MadeOf                    & 2,715       & 2,936       & HasA                    & 2,962       & 3,908       \\ \hline
HasProperty               & 3,573       & 4,885       & Causes                  & 3,705       & 4,737       \\ \hline
Desires                   & 4,121       & 4,185       & HasPrerequisite         & 4,194       & 4,837       \\ \hline
NotDesires                & 4,263       & 4,186       & AtLocation              & 4,497       & 5,983       \\ \hline
SimilarTo                 & 6,980       & 10,676      & PartOf                  & 7,048       & 9,507       \\ \hline
DistinctFrom              & 10,529      & 16,428      & HasSubevent             & 11,899      & 12,969      \\ \hline
MotivatedByGoal           & 11,996      & 12,186      & UsedFor                 & 13,212      & 15,789      \\ \hline
EtymologicallyDerivedFrom & 46,451      & 78,335      & SymbolOf                & 63,785      & 51,298      \\ \hline
DerivedFrom               & 93,190      & 158,921     & EtymologicallyRelatedTo & 97,124      & 145,853     \\ \hline
IsA                       & 100,451     & 127,922     & HasContext              & 133,035     & 135,211     \\ \hline
FormOf                    & 630,914     & 912,022     & Synonym                 & 1,101,134   & 1,356,240   \\ \hline
RelatedTo                 & 1,501,359   & 1,536,157   &                            &             &             \\ \hline
\end{tabular}
\caption{The numbers of triples and entities corresponding to each of the 49 relations in the ConceptNet sample studied in this article. Note that an entity could occur in multiple triples, each with a different relation. 
}
    \label{table: statistics of triple and entity}
\end{table}

We input these 4 million triples into the PBG algorithm for representation learning. We partition the 4 million sampled triples into training, validation and test datasets, containing 75\%, 12.5\%, 12.5\% of the total triples, respectively.  Before doing the sampling, we remove triples with the `ExternalURL' relation. \emph{ExternalURL}  is a `non-semantic' relation that is only referring to a URL identifier and cannot be used for structural analysis of the kind proposed in this paper. Finally, we train and validate PBG on a single server in the Amazon cloud with 4 Intel Xeon cores, with one socket and 61 GB of RAM. After training is concluded, the algorithm outputs a single vector for each unique relation and entity in the training dataset. In the next section, we discuss the validation of the quality of these embeddings.

\subsection{Validating Quality of Embeddings}

Due to the sampling described earlier, a reasonable question arises as to whether the \emph{quality} of the learned representations or `embeddings' output by PBG can be trusted. We propose and use a quantitative measure to validate the quality and effectiveness of these embeddings. Specifically, we first compute a \emph{centroid vector} for each relation, as described below. Recall that we denoted the graph using the symbol $G=(V,R,E)$, where $E$ was the set of triples or `edges' in the graph. In a slight abuse of notation, we use the symbol $G_E$ to represent the set $E$ associated with $G$.  

Given a relation $r \in R$, let $G_r \subseteq G_E$ be the subset of triples in $G_E$ with relation $r$. For each such triple $(h,r,t)$ in $G_r$, we define the \emph{translation vector} $\vec{v} = \vec{t} - \vec{h}$, where $\vec{t}$ and $\vec{h}$ are the embeddings output by PBG for entities $t$ and $h$, respectively. The \emph{centroid vector} $\vec{r_c}$ of $r$ is defined simply as the mean of the translation vectors in $G_r$:

\begin{equation}
  \label{eqn:centroid_vector}
  \vec{r_c} = \frac{1}{|G_r|}\sum_{(h,r,t) \in G_r}(\vec{t} - \vec{h})
\end{equation}

Note that this yields two distinct vectors for $r$: the vector `directly' output by the graph embedding (denoted as $\vec{r}$) and the centroid vector $\vec{r_c}$. We use the symbol $\mathcal{R}$ to indicate the set of directly output embeddings for all 49 relations and the symbol $\mathcal{R}_c$ to indicate the set of (derived) centroid vectors.

With this technical machinery in place, we validate our 4 million-triples sample as follows. First, we calculate two \emph{similarity lists}, $SL_r$ and $SL_r^{'}$, per relation, using each of these two notions of embedding a relation. Specifically, $SL_r$ is constructed as a list of the cosine similarities between $\vec{r}$ and \emph{each} translation vector\footnote{Note that, unlike $\vec{r}$, the translation vector (defined earlier as $\vec{t} - \vec{h}$) clearly depends on the triple.} in $G_r$. The number of entries in $SL_r$ will equal $|G_r|$.  Similarly, $SL_r^{'}$ is constructed as a list of cosine similarities between the centroid vector $\vec{r_c}$ and each translation vector, and also has size $|G_r|$. Furthermore, if we impose an arbitrary ordering on the triples in $G_r$ the two similarity lists are \emph{aligned} by virtue of the common translation vectors computed over triples in $G_r$. However, in the general case, the values in $SL_r$ and $SL_r^{'}$ will differ since the former relies on the direct embedding of $r$ in its construction, while the latter relies on the centroid vector $\vec{r}_c$. 


Given these two per-relation lists, we establish that the two lists are, in fact, highly correlated. The Spearman's rank correlation is designed to measure both the strength and direction of association between two ranked variables and ranges from -1 (perfect negative correlation) to 1 (perfect positive correlation). Because of the geometric features of the embedding space, we are interested in \emph{whether} there is correlation (i.e. the strength), rather than the direction of the correlation. For this reason, given the two aligned lists per relation ($SL_r$ and $SL_r^{'}$), we computed the \emph{absolute value} of the Spearman's rank correlation for each relation, in Table \ref{table: speamans's score}. As expected, some of the correlations are indeed negative. Specifically, of the 49 relations, 24 relations  have a Spearman's rank correlation greater than 0.6, while 25 other relations have negative correlations (approximately 50\%, as would be statistically expected). However, in no case is the absolute value less than 0.4. 

Since $\vec{r_c}$ is a function of the entities in the triples, and never uses the direct embedding $\vec{r}$ output by PBG, this result serves as an independent check on the quality of the embeddings. The high absolute correlations show that, not only are the embeddings learned on our sample self-consistent, but also that they conform closely to the notion of translation that is an important feature of neural graph embeddings \cite{transE}. In contrast, if $\vec{r}$ had showed little or no correlation (compared to $r_c$), it would have begged the question about whether the embeddings were learned by PBG in a sufficiently non-random way that, at least approximately, model the translation operation in vector space. Furthermore, to ensure the results are not an artifact of using Spearman's correlation, we replicated it using an alternate measure (KL-Divergence), with similar conclusions. That is, the distributions of $vec{r_c}$ and $r_c$ were found to exhibit low KL-Divergence for all relations\footnote{As the conclusions are largely identical, we do not reproduce the KL-Divergence table herein.}.  

\begin{table}[!h]
    \centering
    \tiny
    \begin{tabular}{|c|p{1.4cm}|c|p{1.4cm}|}
        \hline
\textbf{Relation}                     & \textbf{Spearman's correlation} & \textbf{Relation}                   & \textbf{Spearman's correlation} \\ \hline
IsA                       & -0.773                 & NotDesires              & 0.954                  \\ \hline
dbpedia/knownFor          & 0.795                  & PartOf                  & -0.939                 \\ \hline
HasSubevent               & 0.882                  & dbpedia/genus           & -0.962                 \\ \hline
Entails                   & -0.958                 & EtymologicallyRelatedTo & -0.385                 \\ \hline
DerivedFrom               & -0.864                 & HasA                    & 0.891                  \\ \hline
UsedFor                   & 0.926                  & Desires                 & 0.946                  \\ \hline
CapableOf                 & 0.934                  & dbpedia/leader          & 0.705                  \\ \hline
AtLocation                & 0.600                  & CreatedBy               & 0.780                  \\ \hline
HasContext                & -0.516                 & NotUsedFor              & 0.639                  \\ \hline
Antonym                   & -0.856                 & DefinedAs               & 0.812                  \\ \hline
HasLastSubevent           & 0.918                  & SymbolOf                & 0.861                  \\ \hline
CausesDesire              & -0.946                 & LocatedNear             & -0.951                 \\ \hline
EtymologicallyDerivedFrom & -0.865                 & HasPrerequisite         & 0.797                  \\ \hline
InstanceOf                & -0.947                 & MadeOf                  & 0.921                  \\ \hline
dbpedia/influencedBy      & -0.475                 & ReceivesAction          & 0.979                  \\ \hline
MannerOf                  & -0.979                 & dbpedia/capital         & 0.946                  \\ \hline
dbpedia/language          & -0.595                 & Causes                  & 0.987                  \\ \hline
HasProperty               & 0.924                  & NotHasProperty          & -0.736                 \\ \hline
dbpedia/product           & -0.880                 & NotCapableOf            & -0.598                 \\ \hline
HasFirstSubevent          & 0.818                  & dbpedia/field           & -0.611                 \\ \hline
dbpedia/genre             & -0.983                 & SimilarTo               & -0.918                 \\ \hline
DistinctFrom              & 0.756                  & MotivatedByGoal         & 0.957                  \\ \hline
dbpedia/occupation        & -0.591                 & ObstructedBy            & 0.849                  \\ \hline
FormOf                    & -0.708                 & RelatedTo               & -0.937                 \\ \hline
Synonym                   & -0.738                 &                            &                        \\ \hline
    \end{tabular}
    \caption{The Spearman's rank correlation score between $SL_r$ and $SL_r^{'}$, for each of the 49 relations. The methodology for constructing these two (aligned) similarity lists is described in the text.}
\label{table: speamans's score}
\end{table}


\subsection{Vectorizing and Clustering Relation-Specific Triples}\label{vectorizing}

Certain relations in ConceptNet are deliberately designed to be broad. A good example is the \emph{HasContext} relation, which is defined on the ConceptNet website as: \emph{A HasContext B} is declared in the knowledge base if `A is a word used in the context of B, which could be a topic area, technical field, or regional dialect'. In this article, we investigate the hypothesis that, despite being originally defined so broadly, there is considerable \emph{substructure} in such relations. In considering the definition of HasContext above, multiple contexts are suggested e.g., \emph{technical field, regional dialect}, and presumably, other contexts that may be similar to these explicit cases. Another example is a relation such as FormOf, where a triple \emph{A FormOf B} may be declared if `A is an inflected form of B; B is the root word of A'. Even the basic official definition \emph{suggests} breadth, since A could either be an `inflected' form of B, or the `root word' of B. Furthermore, there is nothing in the definition that places a strict constraint on such triples, either in theory or in practice. 

Since ConceptNet is crowdsourced to a great extent, it is quite likely that many people have interpreted these relations at `face value' i.e., in accordance with what one would understand their `everyday' meaning to be. Therefore, our goal is to measure and describe the \emph{empirical} substructures, if any, in these three specific relations (\emph{HasContext, FormOf} and \emph{SymbolOf}\footnote{The \emph{SymbolOf} relation is succinctly defined as: the triple \emph{A SymbolOf B} is asserted in the knowledge base if `A symbolically represents B'.}) using a systematic methodology.  An important aspect of these three relations is not just that they are defined broadly and are \emph{coarse-grained}, but are also relatively \emph{high-volume}. Within our sample of 4 million triples, HasContext, FormOf and SymbolOf are asserted in 133,038, 630,914 and 63,785 triples respectively. This provides an added incentive to study these relations further, since they are clearly central to the knowledge base and its purpose of capturing commonsense knowledge as sets of assertions. While these are not the most voluminous relations\footnote{For example, \emph{RelatedTo} and \emph{Synonym} have more than a million triples each, the reason for their breadth (and high volume) is more evident than for a relation such as \emph{HasContext} (for example), since context is a much more ambiguous concept in commonsense reasoning. We hypothesize that a relation like \emph{Synonym} will behave similarly as a relation like \emph{SymbolOf}, although we leave for future work to investigate it. }, we aimed for a set of three relations that are expected to have different practices around them. We leave for future work to replicate our methodology for other such high-volume relations.

An established unsupervised methodology for discovering structure in large collections of data points is \emph{clustering} \cite{clustering}. The relations in ConceptNet were meant to capture common, informative patterns from various data sources that feed into ConceptNet (along with crowdsourcing). If well-defined clusters exist, there is good evidence to suggest that these coarse-grained relations could be further sub-divided or ontologized (possibly by declaring relation-subtypes). By studying both the consistency of the clusters, as well as the subjective nature of data within them, we can start gaining insight into each relation. These insights allow us to gain an empirical understanding of concepts, such as `context' and `form', that are important in commonsense reasoning and communication, beyond their theoretically broad definitions.   


Most established clustering algorithms require the collection and representation of data points to be described in advance. In our case, the goal is to cluster asserted triples of the form $(h,r,t)$ in three independent experiments (with $r$ belonging to one of HasContext, FormOf and SymbolOf in each experiment). However, such a clustering would require us to represent each triple as a vector.  If the relation is fixed, as it would be within an experiment, we can represent the triple using translation vector $\vec{t}-\vec{h}$ that we earlier introduced, and with the entity embeddings $\vec{t}$ and $\vec{h}$ output by PBG. 

For the clustering algorithm itself, we chose to use the classic k-Means algorithm \cite{k-means}. There were several reasons, including the large numbers of data points (which requires efficient clustering), the lack of a task-specific objective function or training labels, and importantly, the methodological preference for an established and reasonably robust clustering algorithm. 

To briefly review k-Means, the algorithm works iteratively to \emph{partition} the dataset into $k$ clusters, each of which is disjoint, owing to the clusters constituting a partition. Let us assume a set $D=\{d_i,\ldots , d_n\}$ of $n$ data points, each of which is $q-$dimensional. We set up the algorithm so that the $k$ means or clusters are randomly initialized, and each of the $n$ data points are assigned to exactly one of the $k$ means, depending on which cluster the point is closest to. Next, the mean for each cluster is re-computed by taking the mean of the vectors assigned to that cluster. The steps above are then repeated: each of the $n$ data points is re-assigned to exactly one of the $k$ clusters (ties are broken arbitrarily), depending on which cluster's mean it is closest to. The means are then re-computed, and so on. We run the algorithm till convergence is achieved, and cluster-assignment of points does not change from one iteration to the next. 

Note that $k$ is a hyper-parameter that must be predefined prior to executing the algorithm. There are several ways to obtain the `best' value of $k$ given a collection of points. The underlying commonality between these methods is to compute, for each value of $k$, an error `score', with lower values implying better quality. This score is computed from the clusters obtained after executing k-Means for that $k$. In practice, $k$ is varied over a predetermined range. By plotting the error score versus $k$, and looking for sudden shifts in the \emph{second derivative} of the curve\footnote{Although the curve can be monotonic for some methods, it is not always guaranteed. Hence, it is incorrect to look for a `minimum'.}, we can determine a value of $k$ that captures the structure in the data. Intuitively, we are seeking a clear `bend' in the \emph{score vs. k} curve to deduce where the second derivative is being minimized.  

\begin{figure}[htbp]
  \begin{minipage}[t]{0.31\textwidth}
    \centering
    \includegraphics[width=1.7in, height=2.0in]{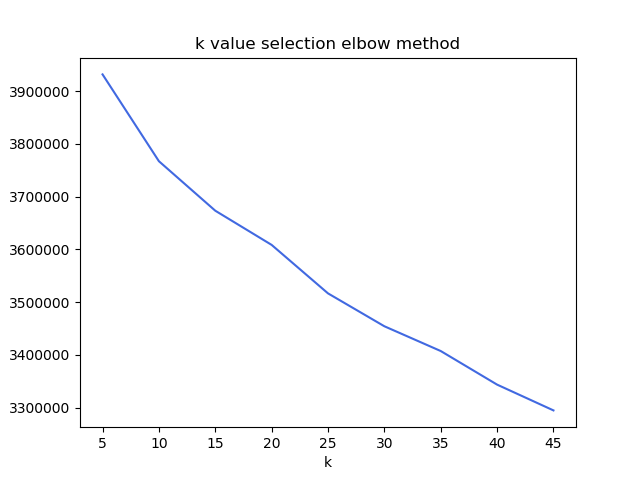}
  \end{minipage}
  \begin{minipage}[t]{0.31\textwidth}
    \centering
    \includegraphics[width=1.7in, height=2.0in]{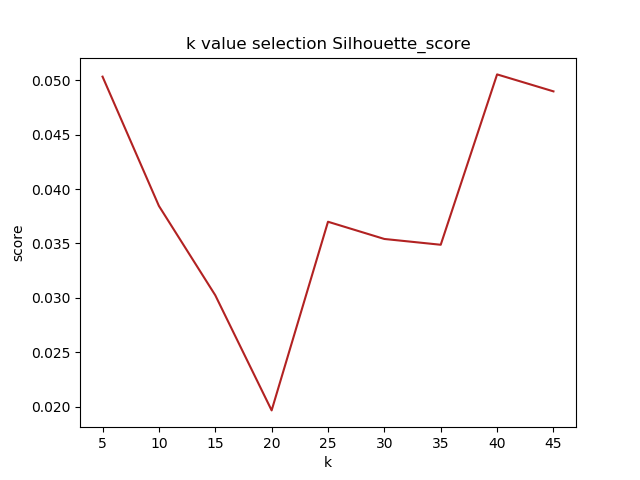}
  \end{minipage}
  \begin{minipage}[t]{0.31\textwidth}
    \centering
    \includegraphics[width=1.7in, height=2.0in]{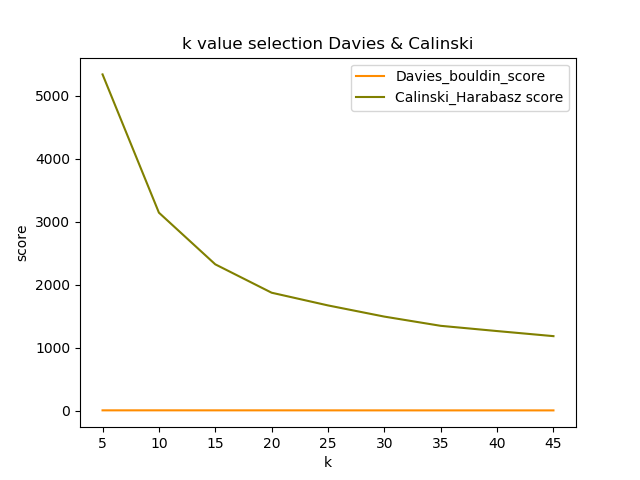}
  \end{minipage}
  \caption{Selection of the parameter $k$ for k-means clustering of HasContext triples, with $k$ on the x-axis and the clustering score on the y-axis, as measured (from left to right) by the elbow method, Silhouette Coefficient, Davies-Bouldin index, and Calinski-Harabasz index, respectively. The last two  are in the same subplot.}
    \label{figure:hascontext-para}
\end{figure} 

A specific method that implements the $k$-selection principles conceptually described above is the \emph{elbow method} \cite{elbow}. The elbow method computes an error-based score based on the `dispersion' of points within each cluster. However, as the first subplot in Figure \ref{figure:hascontext-para} shows, there is no visible decline in the second derivative with $k$. Namely, the `bend', if it even exists, is far too subtle to be useful as a decision-making criterion. For this reason, and also for purposes of robustly selecting a $k$, we also considered three other established alternatives -- the Silhouette Coefficient method, the Davies-Bouldin Index, and the Calinski-Harabasz Index  \cite{SC}, \cite{cluster_measure}. The Silhouette Coefficient value measures how similar a point is to its own cluster's centroid (cohesion) compared to other clusters' centroids (separation). The Index-based measures compute their scores in slightly different ways, but with the same underlying philosophy that clusters should be cohesive and well-separated. Specific details and formulae may be found in the cited works. 

As indicated in Figure \ref{figure:hascontext-para}, while both the elbow method and Davies-Bouldin index are inconclusive, the other two are fairly evident and consistent. In the case of the Silhouette Score, we find that there is an explicit minima at $k=20$. In general, we found $k=20$ to be a robust choice for all three relations (equivalent plots for SymbolOf and FormOf may be found in the supplementary information). The exercise illustrates the methodological utility of using more than one $k$-selection procedure. 

In future work, it may be possible to address the limitation of selecting $k$ heuristically by using hierarchical or agglomerative clustering, and also by using recent clustering algorithms that do not require such hyperparameter selection (e.g., the recent work by \cite{r2clustering}). Our main reason for using k-Means (along with hyperparameter selection methods such as elbow and Calinski-Harabasz Index) for this paper is that it is an established and scalable procedure that can be easily replicated without re-implementation or significant hyperparameter tuning. It also provides a reference and benchmark for future research that is looking to explore the use of other, more advanced algorithms for discovering finer-grained sub-structures in the ConceptNet relations (both the ones that are used in this paper, as well as others, such as RelatedTo, that are not considered in this paper).

\section{Results}

With the selected value of $k=20$ in place, we conducted k-Means clustering for each of the three relations, as discussed earlier. Following the clustering, we computed quantitative metrics to measure the \emph{cohesion} and \emph{separation} of the clusters.  As its name suggests, \emph{cohesion} measures the extent to which the points in each cluster are tightly grouped together. In other words, a cluster with high dispersion has low cohesion. While this intuitive measure can be quantitatively measured in several ways, we consider a simple, easily interpretable methodology and formulae described below in detail. 

First, for each cluster, we compute its centroid and then normalize all points in the clusters, as well as the computed centroid of the cluster\footnote{Since the graph embeddings are not normalized to lie on a unit-radius hypersphere, we normalize the embeddings before computing the distance to enable cross-cluster comparisons, as well as comparisons with the (subsequently described) separation measures.}. Specifically, let us assume $k$ clusters (with $k$ set to 20, as detailed earlier) $\{C_1,\ldots , C_k\}$, which are non-overlapping, non-empty and form a partition over the set $D$ of data points (with each data point $d \in D$ being a vector $\vec{d}=[d_1,\ldots ,d_q]$ with $q$ dimensions, as discussed in Section \ref{vectorizing}) being clustered. The centroid $\vec{c_m}$of a cluster $C_m$ is defined using the formula below:

\begin{equation}
\vec{c_m} = \frac{\sum_{\vec{d_i} \in C_m} \vec{d_i}}{|C_m|}
\end{equation} 

The centroid always exists, since each cluster is non-empty. The sum in the numerator is element-wise. Following centroid computation for each cluster, we normalize each vector $\vec{d}$ in $D$ as well as all $k$ centroids (which are also vectors, with the same dimensionality $q$ as $\vec{d}$), such that the vector now lies on the unit hypersphere, we use the following formula:

\begin{equation}
\vec{v'} = \frac{\vec{v}}{\sqrt{\sum_{i=1}^{i=q}|v[i]|^2}}
\end{equation} 

Here, $\vec{v}$ is any vector from $D$ or a centroid vector, and $\vec{v'}$ is the normalized vector. The division is again element-wise, and $v[i]$ is the $i^{th}$ element of the vector $\vec{v}$. 

Next, we calculate the average \emph{Euclidean} distance (designated as $t_m$) between the normalized points in the cluster $C_m$ and the cluster's centroid $\vec{c_m}$ (which is also normalized):

\begin{equation}
t_m = \frac{\sum_{\vec{d_i} \in C_m} Euc[\vec{d_i},\vec{c_m}]}{|C_m|}
\end{equation} 

Here, $Euc[\vec{x}, \vec{y}]$ between two $q$-dimensional vectors $\vec{x}$ and $\vec{y}$ is $Euc[\vec{x}, \vec{y}]=\sqrt{\sum_{i=1}^{i=q} (x_i - y_i)^2}$, and is a scalar. Since \emph{smaller} distances indicate \emph{greater} cohesion, we subtract the average from 1 to obtain a cohesion $coh_m=1.0-t_m$ of the cluster $C_m$ on a scale of 0.0 to 1.0, with 1.0 indicating perfect cohesion i.e., all points inside the cluster coincide after normalization). In Table \ref{table: average_cohesion}, we report the cohesion for each of the 20 clusters obtained, for each of the three relations. We also report the means and standard deviations, for each of the three relations being studied. Formally, given the cohesions $coh_1, \ldots$, $coh_{k}$, the mean cohesion $M_{coh}$ is given by the formula $\frac{\sum_{i=1}^{i=k} coh_i}{k}$, and the standard deviation $Std_{coh}$ is $\frac{\sum_{i=1}^{i=k} (coh_i - M_{coh})^2}{k}$.

\begin{table}[h!]
\begin{tabular}{|r|r|r|r|}
\hline
\multicolumn{1}{|l|}{\textbf{Cluster ID}} & \multicolumn{1}{l|}{\textbf{FormOf}} & \multicolumn{1}{l|}{\textbf{HasContext}} & \multicolumn{1}{l|}{\textbf{SymbolOf}} \\ \hline
0                      & 4.517                       & 5.953                           & 5.358                         \\ \hline
1                      & 4.576                       & 5.386                           & 4.815                         \\ \hline
2                      & 4.085                       & 4.661                           & 5.436                         \\ \hline
3                      & 4.254                       & 4.640                           & 5.578                         \\ \hline
4                      & 4.785                       & 5.738                           & 4.703                         \\ \hline
5                      & 4.503                       & 5.555                           & 4.735                         \\ \hline
6                      & 4.660                       & 3.284                           & 3.677                         \\ \hline
7                      & 3.601                       & 5.072                           & 4.854                         \\ \hline
8                      & 3.966                       & 4.051                           & 4.623                         \\ \hline
9                      & 4.547                       & 3.731                           & 4.730                         \\ \hline
10                     & 4.065                       & 4.297                           & 6.319                         \\ \hline
11                     & 4.741                       & 3.371                           & 4.553                         \\ \hline
12                     & 4.214                       & 4.215                           & 4.647                         \\ \hline
13                     & 3.918                       & 5.398                           & 4.031                         \\ \hline
14                     & 5.300                       & 4.331                           & 4.715                         \\ \hline
15                     & 4.451                       & 3.739                           & 4.585                         \\ \hline
16                     & 4.648                       & 5.515                           & 4.443                         \\ \hline
17                     & 4.684                       & 4.987                           & 5.132                         \\ \hline
18                     & 5.151                       & 3.677                           & 4.533                         \\ \hline
19                     & 4.119                       & 3.649                           & 5.149                         \\ \hline
{\bf Mean}             & 4.439                       & 4.562                           & 4.831                         \\ \hline
{\bf Std. Dev.}            & 3.302                       & 13.588                          & 6.084                        \\ \hline
\end{tabular}
\caption{The \emph{cohesion} of FormOf, HasContext and SymbolOf clusters, along with per-relation mean and standard deviation. Note that cluster IDs are assigned arbitrarily and independently across relations, and not `aligned' in any way. The IDs are re-used again in visualizations in Figure \ref{fig:clusters}, as well as in Table \ref{table: HasContext} (for HasContext only) where we provide examples of hasContext triples per cluster.}
\label{table: average_cohesion}
\end{table}

Based on the table, we find that the mean cohesion for FormOf, HasContext and SymbolOf clusters is 4.439, 4.562 and 4.831 respectively. While the mean cohesion scores of clusters in these three relations may seem close in value, their standard deviations exhibit significant differences. The standard deviations of HasContext cluster cohesion scores are generally higher than the standard deviations of the other two relations' cohesion scores. This simple result suggests that HasContext may be more diverse (and hence, more \emph{dispersed} in embedding space) than the other two relations. Furthermore, while the deviation is inversely related to the number of triples corresponding to each relation, it is not linear. Finally, it is important to note that the absolute values here are less meaningful than the values relative to one other.

While cohesion is a good measure for characterize clusters, it is not adequate by itself. An `optimal' cohesion can be obtained by assigning each point to its own cluster (in which case, the point becomes the centroid of the cluster). An additional metric, even after controlling for $k$, is the \emph{separation} of the clusters i.e., how `far apart' the different clusters are in the embedding space. Similar to cohesion, there are multiple mathematical ways to capture this qualitative notion. We employ a simple method that is analogous to the cohesion measure--namely, for a given cluster $C_m$, we compute its \emph{separation} $s_m$ by computing the average Euclidean distance from its centroid $\vec{c_m}$ to each of the \emph{other} $k-1=19$ centroids. For simplicity, let us define the centroid-set $\mathcal{C}=\{\vec{c_1},\ldots ,\vec{c_k}\}$ as the set of (normalized) centroids of all $k=20$ clusters:

\begin{equation}
s_m = \frac{\sum_{c_i \in \mathcal{C}, c_i \neq c_m} Euc[\vec{c_i},\vec{c_m}]}{k-1}
\end{equation} 

Note that a subtraction from 1.0 is not necessary (as was the case for the cohesion computations), since the higher the average Euclidean distances between the centroids, the higher the separation. Table \ref{table: average_separation} reports the results for all three relations, along with the mean and standard deviation. The same formulae apply for the mean and standard deviation as noted earlier for cohesion, the only difference being that we use the separations rather than the cohesions. 

We find that, once again, HasContext has highest average separation (4.622). This further suggests that the `contexts' represented by these clusters are well-separated. The FormOf and SymbolOf clusters obtain similar average separations  of 2.985 and 2.970, respectively. Unlike cohesion, the standard deviation of cluster separation scores is highest for the SymbolOf  relation. 

\begin{table}
\begin{tabular}{|r|r|r|r|}
\hline
\textbf{Cluster ID} & \textbf{FormOf} & \textbf{HasContext} & \textbf{SymbolOf} \\ \hline
0           & 2.563  & 4.644      & 2.421    \\ \hline
1           & 2.911  & 3.624      & 2.597    \\ \hline
2           & 2.969  & 3.943      & 5.611    \\ \hline
3           & 2.866  & 5.811      & 2.661    \\ \hline
4           & 2.681  & 4.044      & 2.750    \\ \hline
5           & 2.686  & 3.943      & 2.607    \\ \hline
6           & 3.376  & 6.016      & 4.862    \\ \hline
7           & 3.105  & 3.564      & 2.454    \\ \hline
8           & 2.952  & 4.345      & 2.245    \\ \hline
9           & 2.658  & 5.738      & 2.348    \\ \hline
10          & 4.092  & 4.323      & 4.151    \\ \hline
11          & 2.536  & 6.412      & 2.585    \\ \hline
12          & 2.831  & 4.620      & 2.381    \\ \hline
13          & 3.235  & 3.727      & 4.697    \\ \hline
14          & 4.055  & 4.889      & 2.269    \\ \hline
15          & 2.568  & 4.935      & 2.339    \\ \hline
16          & 2.719  & 3.827      & 2.940    \\ \hline
17          & 3.169  & 3.958      & 2.613    \\ \hline
18          & 3.307  & 5.083      & 2.392    \\ \hline
19          & 2.422  & 4.985      & 2.480    \\ \hline
{\bf Mean}  & 2.985  & 4.622      & 2.970    \\ \hline
{\bf Std. Dev.} & 4.014  & 13.738     & 18.914   \\ \hline
\end{tabular}
\caption{The \emph{separation} of  FormOf, HasContext and SymbolOf clusters, along with per-relation mean and standard deviation.}
\label{table: average_separation}
\end{table}

In comparing the cohesion and the separation of clusters for all three relations in Tables \ref{table: average_cohesion} and \ref{table: average_separation}, we find that mean separation of HasContext clusters is close to their mean cohesion. In other words, the mean distance from a cluster centroid to a within-cluster data point is similar to the mean distance from that cluster-centroid to other cluster-centroids. The mean separations of FormOf and SymbolOf clusters are lower than the respective mean cohesions, suggesting that substructures in these two relations may be less independent than those in HasContext. 

\begin{figure}[htbp]
    \captionsetup{font={small}}
    \centering
    \includegraphics[width=12cm]{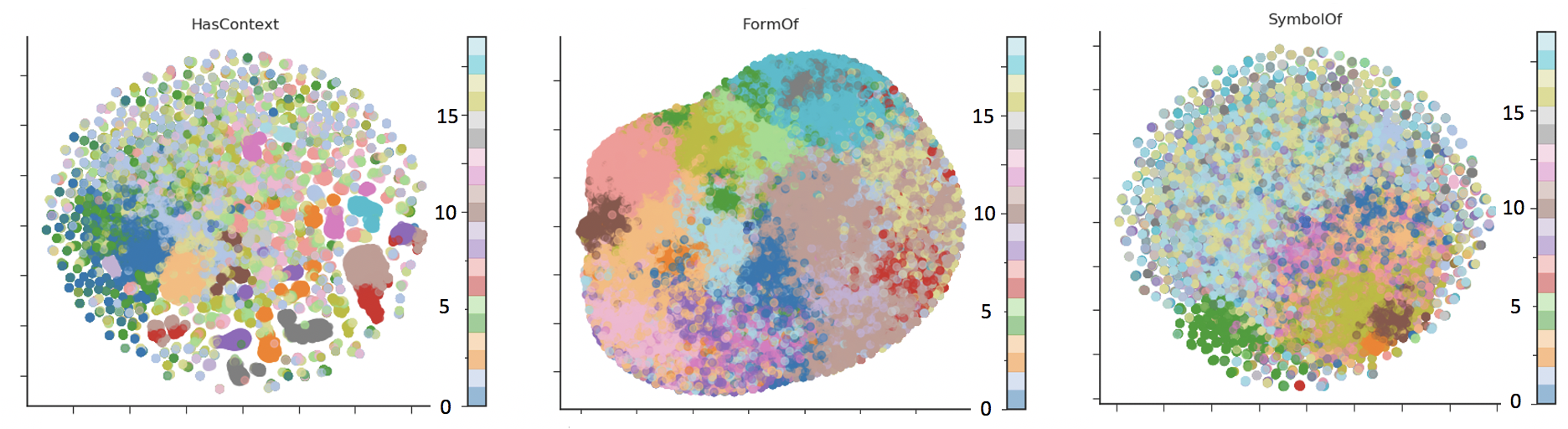}
    \caption{The visualization of HasContext, FormOf and SymbolOf `triples-clusters' using the t-SNE method. Here, 20 such clusters per relation, discovered using k-Means, are represented using different colors. The cluster IDs are indicated on the color bar on the right side of each subplot, and are consistent with those introduced in Tables \ref{table: average_cohesion} and \ref{table: average_separation}, and used subsequently in Table \ref{table: HasContext}. The t-SNE `dimensions' lack intrinsic meaning and are only for visualization purposes. }
    \label{fig:clusters}
  \end{figure}

We can also visualize the clustering results by first performing dimensionality reduction (to two dimensions) using the t-Stochastic Neighbor Embedding (t-SNE) method, which has emerged as a state-of-the-art neural visualization technique in the machine learning community \cite{tsne}. Next, we plot these points in 2D space by using a different color to represent each cluster. Results for all three relations are visualized in Figure \ref{fig:clusters}. For all three relations (and especially, FormOf), there are some homogeneous clusters, where the embeddings are close to each other. However, other clusters can exhibit dispersion. For both SymbolOf and HasContext, some clusters exhibit high dispersion and overlap with other clusters. These dispersed clusters provide an explanation for why HasContext and SymbolOf were found to have much high standard deviations on both the \emph{separation}  and \emph{cohesion} measures described earlier (compared with much lower standard deviations for FormOf).

%
%
%
It is also instructive to study the substructures for a given relation by further analyzing triples \emph{qualitatively} within each of the 20 clusters. We conduct such a qualitative study for the HasContext relation. Specifically, we randomly sample five triples per cluster to determine if we can deduce the `theme' of the cluster from these five triples. These samples are reproduced in Table \ref{table: HasContext}.

Even the limited set of sampled triples (per cluster) in the table demonstrates a pattern. For example, some of the clusters clearly embody scientific `domains' of study such as geography (Cluster 14), chemistry (Cluster 6), medicine (Cluster 17), mathematics (Cluster 18), and physics (Cluster 19). However, there are also `mixed' clusters that seem to be related to more than one theme, at least on the surface. Cluster 3 contains some `locality' triples, even though Cluster 15 is predominantly concerned with localities, and Cluster 3 is mainly concerned with zoology. 

We believe that there could be two causes or interpretations for such `confusion'. The first is due to the automatic and unsupervised nature of the embedding and the second is due to ConceptNet itself, both in terms of the noise within ConceptNet and also because some tail entities, such as \emph{/c/fr/localites}, may be imbalanced in terms of the head entities associated with them. Some other clusters also include some interesting combinations, such as Cluster 12 which contains triples corresponding to both `computing' and `slang'. In the embedding and clustering process, `computing' and `slang'-related triples are thought to be close to each other in vector space, although their semantic similarity is not intuitively evident. Yet other clusters, such as Cluster 0, seem to encapsulate the broad notion of \emph{HasContext}, and do not have an evident thematic classification that we can determine. 

Additional interpretations of these overlapping clusters are also possible. One interpretation is \emph{historical context}, especially concerning how these fields (`sub-structures') have co-evolved over time. For example, fields such as mathematics, physics and even computing have become increasingly entwined over time. The overlap may simply be a consequence of this co-evolution. Another possibility is that the fields share common semantics (including common terms), and this is reflected in overlap as well. Finally, we note that ConceptNet is meant to be a repository of commonsense knowledge, and not necessarily a detailed description of highly specialized domains within science (or other areas). The overlap may be arising not because the fields themselves are highly overlapping but because their \emph{commonsense content} is. We leave for future work to investigate these interpretations more rigorously and quantitatively. 

Aside from hypothesis validation of commonsense knowledge and evolution of commonsense semantics in specialized areas, the sub-structures could be used in novel \emph{domain-specific} applications of AI, including domain-specific versions of fuzzy clustering and expert systems \cite{fuzzy1}, \cite{expertsystems1}, \cite{expertsystems2}. One such application is in knowledge acquisition: our discovered sub-structures could be used for isolating specific portions of, or triples within, ConceptNet that are useful for the application domain being considered. Once isolated, these triples could be used in a domain-specific fuzzy clustering algorithm or expert system to enhance performance. Further research is needed to build and evaluate such applications. 

\begin{center}\tiny
\begin{longtable}[h!] {|c|c|c|}
\caption{Examples of five triples randomly sampled from each of the 20 \emph{HasContext} clusters. The cluster IDs are consistent with those used in Figure \ref{fig:clusters} (left figure). Columns 2 and 3 list the head and tail entities respectively, with the relation always being \emph{hasContext}, by definition.}\label{table: HasContext} \\ \hline

%
%
%
\multirow{5}{*}{0}  & /c/en/immunosenescent/a   & /c/en/pathology                       \\ \cline{2-3} 
                    & /c/en/handball/v          & /c/en/soccer                          \\ \cline{2-3} 
                    & /c/en/screenplay/n        & /c/fr/cinema                         \\ \cline{2-3} 
                    & /c/en/crivvens            & /c/en/scotland                        \\ \cline{2-3} 
                    & /c/en/rhizomatic/a        & /c/en/philosophy                      \\ \hline
\multirow{5}{*}{1}  & /c/fr/sapide/a        & /c/en/literary                        \\ \cline{2-3} 
                     & /c/hu/szirn/n               & /c/en/literary                        \\ \cline{2-3} 
                    & /c/ga/eo/n/wikt/en\_3     & /c/en/literary                        \\ \cline{2-3} 
                    & /c/af/elk/n               & /c/en/literary                        \\ \cline{2-3} 
                    & /c/ga/gair/v/wikt/en\_1   & /c/en/literary                        \\ \hline
\multirow{5}{*}{2}  & /c/it/vena\_cava/n        & /c/en/anatomy                         \\ \cline{2-3} 
                    & /c/et/fluor/n             & /c/fr/chimie                          \\ \cline{2-3} 
                    & /c/fr/saksaoul/n          & /c/fr/botanique                       \\ \cline{2-3} 
                    & /c/mul/raw                & /c/fr/linguistique                    \\ \cline{2-3} 
                    & /c/fr/tagbanoua/n         & /c/fr/linguistique                    \\ \hline
\multirow{5}{*}{3}  & /c/fr/schipluiden/n       & /c/fr/localites                      \\ \cline{2-3} 
                    & /c/en/brontotherid/n      & /c/en/zoology                         \\ \cline{2-3} 
                    & /c/fr/de\_hem/n           & /c/fr/localites                       \\ \cline{2-3} 
                    & /c/fr/brozolo/n           & /c/fr/localites                       \\ \cline{2-3} 
                    & /c/en/onychoteuthid/n     & /c/en/zoology                         \\ \hline
\multirow{5}{*}{4}  & /c/ha/umra/n              & /c/en/islam                           \\ \cline{2-3} 
                    & /c/no/jordakse/n          & /c/en/geometry                        \\ \cline{2-3} 
                    & /c/en/tsar/n              & /c/en/historical                      \\ \cline{2-3} 
                    & /c/lij/dısette            & /c/en/cardinal                        \\ \cline{2-3} 
                    & /c/mi/iwa                 & /c/en/cardinal                        \\ \hline
\multirow{5}{*}{5}  & /c/fr/scheelite/n         & /c/en/mineral                         \\ \cline{2-3} 
                    & /c/de/natriumdichromat/n  & /c/en/inorganic\_compound             \\ \cline{2-3} 
                    & /c/en/oxazepane/n         & /c/en/organic\_compound               \\ \cline{2-3} 
                    & /c/en/gelsemine/n         & /c/en/organic\_compound               \\ \cline{2-3} 
                    & /c/en/conductin/n         & /c/en/protein                         \\ \hline
\multirow{5}{*}{6}  & /c/en/azodicarbonamide/n  & /c/en/chemistry                       \\ \cline{2-3} 
                    & /c/en/ricinoleate/n       & /c/en/chemistry                       \\ \cline{2-3} 
                    & /c/fi/rikkiyhdiste/n      & /c/en/chemistry                       \\ \cline{2-3} 
                    & /c/en/test/v/wikt/en\_1   & /c/en/chemistry                       \\ \cline{2-3} 
                    & /c/en/vinyl\_acetate/n    & /c/en/chemistry                       \\ \hline
\multirow{5}{*}{7}  & /c/en/business/n          & /c/en/los\_angeles                    \\ \cline{2-3} 
                    & /c/sq/shkret??roj/v        & /c/en/tosk                            \\ \cline{2-3} 
                    & /c/en/hooklet/n           & /c/en/natural\_history                \\ \cline{2-3} 
                    & /c/da/femten                 & /c/en/cardinal            \\ \cline{2-3} 
                    & /c/it/un/a                & /c/en/sometimes\_before\_vowel\_or\_h \\ \hline
\multirow{5}{*}{8}  & /c/fr/deontologie/n       & /c/en/philosophy                      \\ \cline{2-3} 
                    & /c/en/cap\_cloud/n        & /c/en/meteorology                     \\ \cline{2-3} 
                    & /c/en/back\_ganging/n     & /c/en/uk                              \\ \cline{2-3} 
                    & /c/en/syringic/a          & /c/en/organic\_chemistry              \\ \cline{2-3} 
                    & /c/en/diethenoid/a        & /c/en/organic\_chemistry              \\ \hline
\multirow{5}{*}{9}  & /c/en/meteor/n            & /c/en/martial\_arts                   \\ \cline{2-3} 
                    & /c/fr/pause/n             & /c/fr/musique                         \\ \cline{2-3} 
                    & /c/cs/moderator/n         & /c/en/uk                              \\ \cline{2-3} 
                    & /c/en/lin/v/wikt/en\_1    & /c/en/uk                              \\ \cline{2-3} 
                    & /c/ms/kata\_benda/n       & /c/en/grammar                         \\ \hline
\multirow{5}{*}{10} & /c/en/neurodegeneration/n & /c/fr/neurologie                      \\ \cline{2-3} 
                    & /c/no/oppholde/v          & /c/en/somewhere                       \\ \cline{2-3} 
                    & /c/scn/lu                 & /c/en/definite\_article               \\ \cline{2-3} 
                    & /c/en/monotypy/n          & /c/en/conservation\_biology           \\ \cline{2-3} 
                     & /c/nl/wao/n     & /c/en/netherlands                        \\ \hline
\multirow{5}{*}{11} & /c/sl/kriptozoologija/n   & /c/fr/biologie                        \\ \cline{2-3} 
                    & /c/nl/neptunus/n          & /c/fr/astronomie                      \\ \cline{2-3} 
                    & /c/fr/l1/n/wikt/fr\_2     & /c/fr/astronomie                      \\ \cline{2-3} 
                    & /c/en/freedom\_rider/n    & /c/en/politics                        \\ \cline{2-3} 
                    & /c/fr/corps/n             & /c/fr/numismatique                    \\ \hline
\multirow{5}{*}{12} & /c/en/vamptastic/a        & /c/en/slang                           \\ \cline{2-3} 
                    & /c/de/funzen/v/wikt/en\_1 & /c/en/slang                           \\ \cline{2-3} 
                    & /c/fi/filu/n              & /c/en/computing                       \\ \cline{2-3} 
                    & /c/en/dep/n               & /c/en/computing                       \\ \cline{2-3} 
                    & /c/en/nonserver/a         & /c/en/computing                       \\ \hline
\multirow{5}{*}{13} & /c/en/fiscal/n/wikt/en\_1 & /c/en/legal                           \\ \cline{2-3} 
                    & /c/de/silver\_goal/n      & /c/en/football                        \\ \cline{2-3} 
                    & /c/en/shitcan/v           & /c/en/vulgar                          \\ \cline{2-3} 
                    & /c/en/eicosanoid/n        & /c/fr/biochimie                       \\ \cline{2-3} 
                    & /c/rm/mel/n               & /c/en/rumantsch\_grischun             \\ \hline
\multirow{5}{*}{14} & /c/fr/ocean\_atlantique/n & /c/fr/geographie                      \\ \cline{2-3} 
                    & /c/sl/balkanski/a         & /c/fr/geographie                      \\ \cline{2-3} 
                    & /c/fr/riviera/n           & /c/fr/geographie                      \\ \cline{2-3} 
                     & /c/nl/ijsvlakte/n              & /c/fr/geographie                      \\ \cline{2-3} 
                    & /c/fr/sapouy/n            & /c/fr/geographie                      \\ \hline
\multirow{5}{*}{15} & /c/fr/lapedona/n          & /c/fr/localites                      \\ \cline{2-3} 
                    & /c/fr/tour\_de\_faure/n   & /c/fr/localites                       \\ \cline{2-3} 
                    & /c/fr/amendeuix\_oneix/n  & /c/fr/localites                       \\ \cline{2-3} 
                    & /c/fr/espedaillac/n       & /c/fr/localites                       \\ \cline{2-3} 
                    & /c/fr/rye/n               & /c/fr/localites                       \\ \hline
\multirow{5}{*}{16} & /c/fro/voleir/n           & /c/en/anglo\_norman                   \\ \cline{2-3} 
                    & /c/en/stoater/n           & /c/en/horse\_racing                   \\ \cline{2-3} 
                    & /c/nrf/malon/n            & /c/en/jersey                          \\ \cline{2-3} 
                    & /c/en/antieczema/a        & /c/en/pharmacology                    \\ \cline{2-3} 
                    & /c/ga/heitribh?©ascna/n    & /c/en/linguistics                     \\ \hline
\multirow{5}{*}{17} & /c/fr/humoral/a           & /c/fr/medecine                       \\ \cline{2-3} 
                    & /c/en/fasciculatory/a     & /c/en/medecine                        \\ \cline{2-3} 
                    & /c/fr/anti?©pileptique/a   & /c/fr/medecine                       \\ \cline{2-3} 
                    & /c/de/tropf/n             & /c/fr/medecine                        \\ \cline{2-3} 
                    & /c/fr/sida/n/wikt/fr\_1   & /c/fr/medecine                        \\ \hline
\multirow{5}{*}{18} & /c/en/polymodality/n      & /c/en/mathematics                     \\ \cline{2-3} 
                    & /c/en/biplanar/a          & /c/en/mathematics                     \\ \cline{2-3} 
                    & /c/it/esaedro/n           & /c/en/mathematics                     \\ \cline{2-3} 
                    & /c/de/divergieren/v       & /c/en/mathematics                     \\ \cline{2-3} 
                    & /c/en/local\_maximum/n    & /c/en/mathematics                     \\ \hline
\multirow{5}{*}{19} & /c/en/thermoelasticity/n  & /c/en/physics                         \\ \cline{2-3} 
                    & /c/pt/hidrostatico/a      & /c/en/physics                         \\ \cline{2-3} 
                    & /c/en/remanence/n         & /c/en/physics                         \\ \cline{2-3} 
                    & /c/en/specific/a          & /c/en/physics                         \\ \cline{2-3} 
                    & /c/en/microelectronvolt/n & /c/en/physics                         \\ \hline
\end{longtable}
\end{center}

Further investigation of the inter-relationships between these clusters in Figure \ref{fig:clusters} yields other insights. For example, the cluster focused on chemistry-related triples overlaps with the cluster containing biology, as well as with astronomy-related, triples, as we would intuitively expect. While some of the overlap in the figure is exaggerated due to dimensionality reduction, it is nonetheless indicative of the low separation between these two clusters in high-dimensional space. It is an indirect acknowledgement of the shared lineage of these scientific disciplines. An interesting avenue for future exploration is to quantify and explain the observed topical overlap between such clusters, by using techniques such as hierarchical clustering.

\section{Discussion}

In exploring three specific relations (SymbolOf, FormOf, and HasContext), we found and characterized significant `substructures' that are \emph{thematically diverse} (especially in the case of HasContext), illustrating distinct and complex sub-relations within the overall relation. Sub-structures were also noted in FormOf and SymbolOf, but were less interesting and had clear separations than HasContext. 

We have also empirically observed that, while `super-class' semantics tend to be associated with the definition of HasContext\footnote{Namely, when head entity $h$ is a word that is used in the context of tail entity $t$, $t$ tends to be a more general, abstract `super-class' of $h$, such as a topic area, technical field, or regional dialect, as is also mentioned in the official definition of HasContext}, there are significant substructures that can't be uniformly explained by an `umbrella' term like HasContext. These substructures may help us better understand what the different contexts are in which people interpret pairs of words or entities. Understanding context is critical for building systems that have commonsense, such as chatbots and conversational agents, that need to understand sentences in the specific context in  which the sentences are uttered. 

Another interesting case is  SymbolOf. The ratio of the size of the head-entity set of SymbolOf to the size of its tail-entity set is 0.011, which is almost 1/200 of the average value observed for other relations. This potentially implies that the number of words or phrases that are used to `describe' the symbols is far greater than the number of symbols themselves\footnote{One reason for this implication is that the head entity of `SymbolOf' is a symbol and the tail entity is some kind of description, name or other information pertinent to characterizing that symbol.}. Symbols, and arguably, by extension, the emojis used on social media, are necessarily under-determined, and the semantics assigned to symbols vary in different contexts. While not qualitatively unsurprising, our results suggest that the `diffusion' of symbol-semantics is far greater than one might have thought. We leave a detailed study of this diffusion for future research.

\section{Conclusion and Future Work}
Commonsense knowledge and reasoning are not only ubiquitous among humans, they have also been recognized as essential for building `general' AI architectures with human-like reasoning abilities, especially when facing uncertain, implicit, or even potentially contradictory, information. Despite much progress in commonsense-related tasks such as question answering, there is a lack of understanding of structural properties of commonsense knowledge. At the same time, the release and growth of commonsense knowledge graphs, such as ConceptNet, has provided an opportunity to conduct such a study using rigorous and replicable computational techniques.

In this article, we presented and applied such a data-driven methodology to understand structure in commonsense assertions through an empirical study of three high-volume, coarse-grained relations, namely, FormOf, SymbolOf and HasContext. All of these relations (and especially, hasContext) are known to be important in everyday commonsense tasks, including communication and conversation. Using both qualitative and quantitative analyses, we found that there are at least 20 distinct kinds of context that can be discovered within ConceptNet, some very well-defined (such as a scientific field of study), with others being more diffuse. Similar findings seem to apply for the other two relations. In some cases, there are unusual, but non-random, degrees of overlap and association between contexts, such as computer science and slang. Some of these contexts could be used to semi-automatically refine ConceptNet and develop more comprehensive ontological resources for the NLP community. Similarly, different sub-categories of symbols and forms can be semi-automatically discovered.

There are several promising opportunities for future research. For example, it may be worthwhile going even deeper into a relation like HasContext to discover if there are \emph{hierarchical} substructures, rather than a single set of substructures. Hierarchical clustering algorithms in the graph embedding space could be used to achieve this goal methodologically. An even more ambitious line of study would be to connect these empirical results to theoretical claims (e.g., by \cite{csbook1}) about commonsense knowledge. Replicating the methodology on other coarse-grained and high-volume relations, both in ConceptNet and other knowledge bases, is also a valuable avenue for future investigation. 

Finally, while the use of ConceptNet and other static resources is valuable for studying commonsense knowledge at a given time, they may not necessarily account for dynamic changes in meanings and semantics that occur over time. Future research may want to take into account other graph structures that can grow and change over time, such as co-occurrence graphs. From those graphs,  coarse-grained relationships could be  extracted, similar to the approach presented in this paper, and the consequently extracted sub-structures might better reflect topical relationships with respect to a reference time-slice. Additionally, taxonomies can be automatically extracted from such sub-structures. Expanding the analysis in this paper using such co-occurrence graphs, rather than just the sub-graph from a single relation (within the context of a single experiment), is a promising area for future work.

\section*{Declarations}

\subsection*{Funding}
This project was funded under MOWGLI, a project in the Defense Advanced Research Projects Agency (DARPA)
Machine Common Sense program, supported by United States Office Of Naval Research under Contract No.
N660011924033. The views expressed in this paper are solely of the authors.

\subsection*{Conflicts of interest/Competing interests}

The authors declare that they have no competing interests.

\subsection*{Availability of data and material}

The subject of study is ConceptNet 5.7, downloadable from \url{http://www.conceptnet.io/}. Identifiers of the 4 million sampled triples are available at \url{https://drive.google.com/file/d/1RlHkwvuYmgOMqNf4UT0EDx5MmnbKvX4s/view?usp=sharing}.

\subsection*{Code availability}

The study relies on publicly available codebases that have been developed by others, including (i) t-SNE: \url{https://scikit-learn.org/stable/modules/generated/sklearn.manifold.TSNE.html}; (ii) PBG: \url{https://github.com/facebookresearch/PyTorch-BigGraph}, and (iii) standard clustering and machine learning packages available at \url{https://scikit-learn.org/stable/}. 

\subsection*{Authors' contributions}

Kejriwal was responsible for research supervision, project management, formulation of research question, and editing. Shen was responsible for all core research, including experimental work and writing.

\end{document}